\documentclass[letterpaper, 10 pt, conference]{ieeeconf}  
\IEEEoverridecommandlockouts                              

\usepackage{cite}
\usepackage{amsmath,amssymb,amsfonts}
\usepackage{graphicx}
\usepackage{textcomp}
\usepackage{xcolor}
\usepackage{subfigure}
\usepackage{booktabs}
\usepackage{algorithm}
\usepackage{threeparttable}
\usepackage{hyperref}
\usepackage{algpseudocode}
\usepackage[bottom]{footmisc}
\usepackage{multirow}

\title{Improvement on LiDAR-Camera Calibration Using Square Targets
}

\author{Zhongyuan Li\textsuperscript{\textsection}, Honggang Gou\textsuperscript{\textsection}, Ping Li, Jiaotong Guo and Mao Ye\textsuperscript{\textdagger} \\
\thanks{\textsuperscript{\textsection}Equal contribution.}%
\thanks{\textsuperscript{\textdagger}Corresponding author.}%
\thanks{Zhongyuan Li, Honggang Gou, Ping Li, Jiaotong Guo and Mao Ye are with Inceptio Technology, Shanghai, China. \{zhongyuan.li, honggang.gou,ping.li,jiaotong.guo,mao.ye\}@inceptio.ai}
}

\def\BibTeX{{\rm B\kern-.05em{\sc i\kern-.025em b}\kern-.08em
    T\kern-.1667em\lower.7ex\hbox{E}\kern-.125emX}}
\begin{document}

\maketitle
\thispagestyle{empty}
\pagestyle{empty}

\begin{abstract}
Precise sensor calibration is critical for autonomous vehicles as a prerequisite for perception algorithms to function properly. Rotation error of one degree can translate to position error of meters in target object detection at large distance, leading to improper reaction of the system or even safety related issues. Many methods for multi-sensor calibration have been proposed. However, there are very few work that comprehensively consider the challenges of the calibration procedure when applied to factory manufacturing pipeline or after-sales service scenarios. In this work, we introduce a fully automatic LiDAR-camera extrinsic calibration algorithm based on targets that is fast, easy to deploy and robust to sensor noises such as missing data. The core of the method include: (1) an automatic multi-stage LiDAR board detection pipeline using only geometry information with no specific material requirement; (2) a fast coarse extrinsic parameter search mechanism that is robust to initial extrinsic errors; (3) a direct optimization algorithm that is robust to sensor noises. We validate the effectiveness of our methods through experiments on data captured in real world scenarios.
\end{abstract}

\section{Introduction}
Autonomous Driving (AD) systems are commonly equipped with Light Detection and Ranging (LiDAR) sensors and cameras. Precise intrinsic and extrinsic parameters are crucial for data fusion from these sensors, for which a dedicated calibration procedure is required. In real world scenarios, the calibration procedure is needed both during the manufacturing stage as well as in after-sales service when some sensors need to be replaced. While designing a calibration system for mass production manufacturing pipeline and real world after-sales workshops, there are several critical requirements or constraints to be considered as follows.
\begin{itemize}
\item First of all, manufacturing of the vehicles are typically streamlined and only limited amount of time is allowed for each step to ensure production speed. Therefore, the entire calibration procedure should be fast, typically within a few seconds. Towards this end, a fully automatic system without relying on any human intervention is desired. 
\item Secondly, the calibration method should be applicable to different types of LiDAR sensors regardless of the underlying sensing mechanism. 
\item Thirdly, no assumption of structural environments can be made due to large variations in after-sales workshops. Instead, calibration targets are generally acceptable as long as they are easy to fabricate (e.g., without requiring certain particular materials), maintain and deploy.
\item Last but not the least, the calibration should be able to cope with partial LiDAR observation in close up scans or degraded data quality at large distance. 
\end{itemize}

Existing calibration methods do not comprehensively consider all these requirements. Some require human assistance such as selection of feature points\cite{Yechenginteractive2019}. Alternatively, automatic detection of calibration patterns is possible; yet materials that are friendly to LiDAR reflection is normally needed\cite{huang2020improvements}. Methods that require no calibration target assume suitable calibration environments, which are not widely available in after-sales workshops. In this work, we propose a fully automatic method based on calibration targets that fulfills the requirements listed above. The key contributions of our method, which also reflect the core components of the system, are summarized as follows:
\begin{itemize}
\item A fully automatic algorithm for LiDAR-camera calibration based on square targets.
\item A multi-stage LiDAR board detection algorithm relying only on geometric information, applicable to different types of LiDAR sensors (Section ~\ref{LMD}). No particular material is required for the calibration target.
\item A direct optimization method that is more accurate and robust to sparse and incomplete LiDAR observation compared to the state of art method (Section ~\ref{sec:opt}).
\end{itemize}

\section{Related work}
Existing work on extrinsic calibration between a LiDAR and a camera can be roughly categorized according to the type of features used for data association and optimization. Each line of work is briefly summarized in the following. Another way of categorizing existing work is based on the optimization method, leading to learning based methods and non-learning based methods. We dedicate the last part of this section to the learning-based methods.
\subsection{Target based Methods}
Target-based approaches use specially designed objects to facilitate feature detection and matching. In \cite{zhang2004extrinsic}, a checkerboard is used for the calibration of a 2D laser range finder and a camera. Laser points are manually selected and the parameters are estimated using plane-line correspondences. A checkerboard is also utilized in \cite{zhou2018automatic}, where the segmentation of target point clusters in LiDAR is not discussed. The extraction of checkerboard edges in point cloud is affected by the measurement quantization error, especially at large distances, leading to degraded calibration accuracy \cite{huang2020improvements}. In \cite{huang2020improvements}, LiDAR vertices are estimated indirectly by optimizing the transformation between LiDAR and target coordinate frames, where LiDAR targets are detected based on intensities. A good detection performance relies on the target material, target range and the point cloud density, limiting the application scope of the algorithm, as shown in Fig.~\ref{lidar_marker_det_sup_a}. Both \cite{beltran2022automatic} and \cite{yan2023joint} design a rectangular calibration board with four circular holes with different maker patterns. A v-shaped target~\cite{kwak2011extrinsic} has also been explored as the calibration target, where extrinsic parameters are estimated via solving the perspective-n-point (PnP) problem. A sphere target has also been leveraged for calibration in \cite{lee2017calibration}. In this paper, square boards are chosen as the calibration target due to their ease of fabrication, enabling easy deployment of the proposed method in real world scenarios.      
\subsection{Appearance based Methods}
Image and LiDAR intensity map are used within the mutual information framework for extrinsic calibration \cite{pandey2012automatic}. Issues arise when the range sensor does not provide reflectivity information or the observed scene has low reflectivity. In \cite{shin2016computational}, extrinsic parameters are estimated by the optimization of the image intensity variation. This approach assumes constant lighting and suffers from  significantly degraded performance in case of illumination variation.
\subsection{Edge based Methods}
LiDAR-camera calibration can be accomplished via registration of LiDAR depth discontinuities and image edges, either using the mutual information framework \cite{taylor2012mutual}, the grid search method \cite{levinson2013automatic} or the optimization method minimizing line projection error \cite{li2022accurate}. LiDAR edges can be extracted from a single LiDAR scan \cite{levinson2013automatic} or from a point cloud map built using LiDAR odometry algorithms \cite{yuan2021pixel}. A primary limitation of edge based methods is their assumption that depth discontinuities corresponds mostly to image edges which may be not valid in practice. Moreover, these algorithms require edges in various directions to avoid degenerate optimization.   
\subsection{Learning based Methods}
Reg-Net\cite{schneider2017regnet} is the pioneering work using deep learning for LiDAR-camera calibration. The techniques in \cite{schneider2017regnet} are followed by subsequent works, including the automatic generation of training data by adding perturbation to known calibration, the dual branch architecture for camera and LiDAR feature extraction and the training of several networks for iterative calibration refinement. Lcc-Net\cite{lv2021lccnet} is another representative work which uses a cost volume layer for LiDAR-camera feature matching. DXQ-Net \cite{jing2022dxq} learns the calibration flow defined as the offset of the current LiDAR projection in image space. A differentiable optimizer is employed for solving extrinsic parameters. Compared to \cite{lv2021lccnet}, its generalization ability is greatly enhanced with respect to different environments and sensor installation. However, its accuracy and stability declines noticeably when applied to different sensor models. The generalization issue is common to learning based methods. Besides, the model training requires ground truth calibrations which come from other calibration approaches such as the target-based methods.         
\begin{figure}[htbp]
\centering
\subfigure[noisy intensity]{
\label{lidar_marker_det_sup_a}
\includegraphics[scale=0.3]{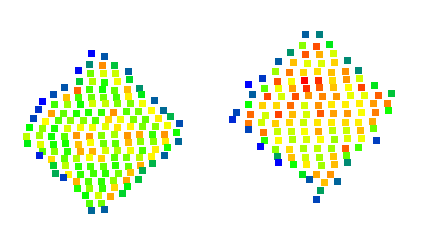}
}
\subfigure[simulated board]{
\label{lidar_marker_det_sup_b}
\includegraphics[scale=0.15]{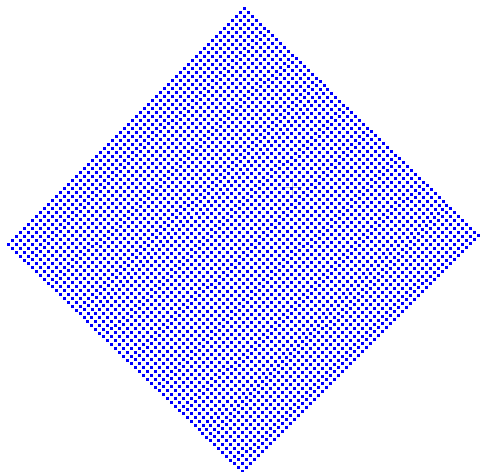}
}
\caption{(a) shows that the marker pattern is difficult to identify from LiDAR intensity information. (b) is simulated LiDAR board points for generating the reference descriptor for LiDAR marker detection.}
\label{lidar_marker_det_sup}
\end{figure}
\section{Methodology}\label{Metho}
\subsection{System Overview}
\begin{figure}[htbp]
	\centering
    \includegraphics[scale=0.2]{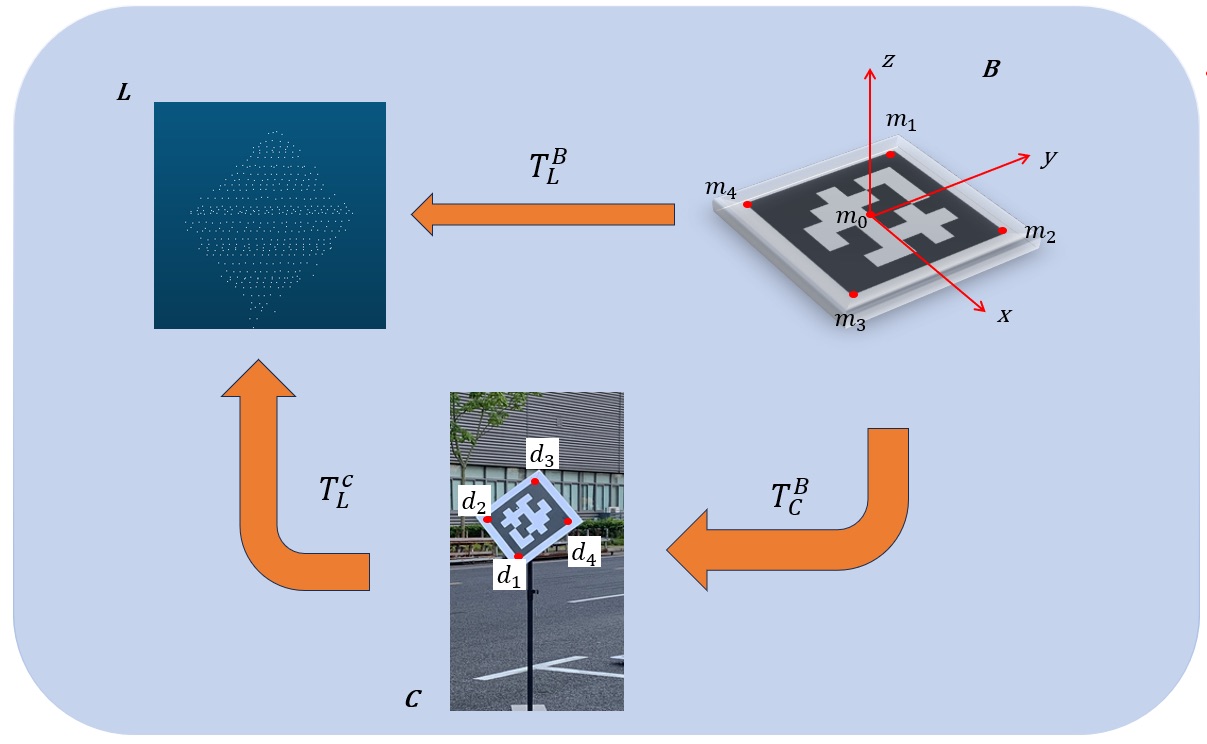}
	\caption{Coordinate System Definition}
	\label{FIG:1}
\end{figure}

\begin{figure*}
	\centering
    \includegraphics[scale=0.4]{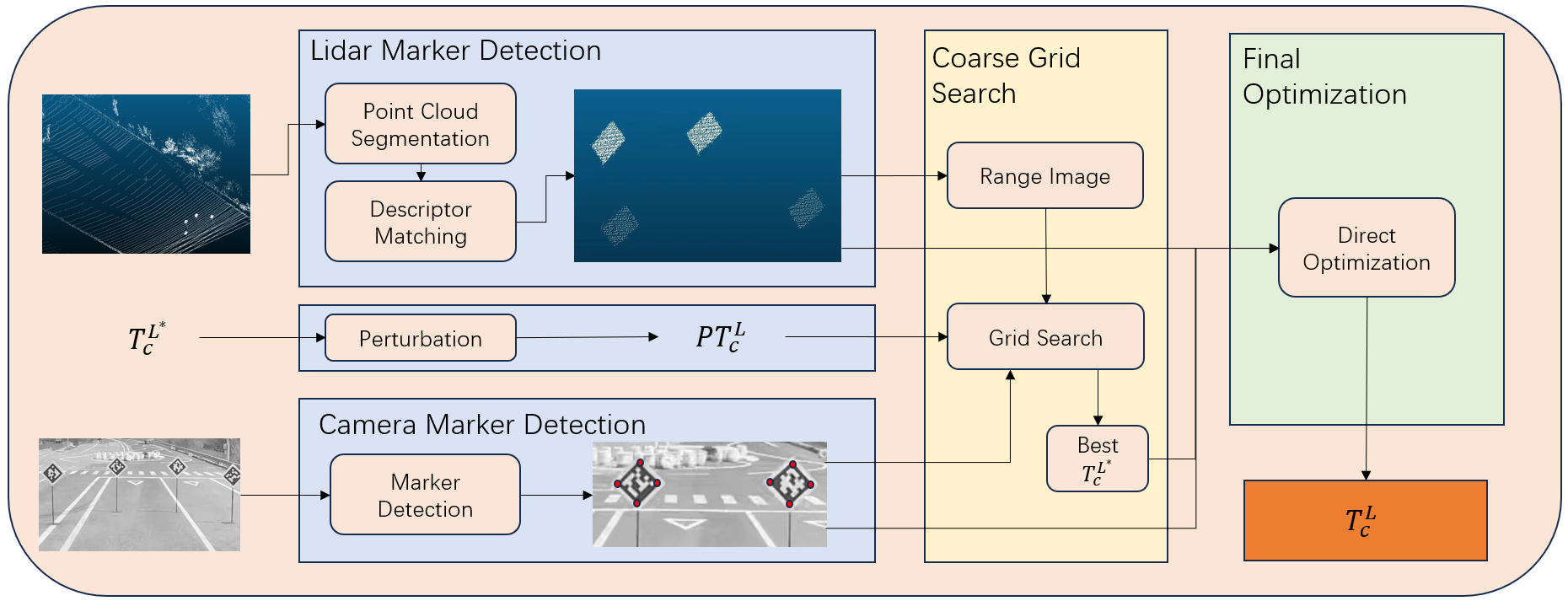}
	\caption{The proposed fully automatic pipeline for LiDAR-camera extrinsic calibration.}
	\label{FIG:2}
\end{figure*}

A rigid transformation between two different coordinate systems can be represented by a matrix $\mathit{T} = \begin{bmatrix} \mathit{R} & \mathit{t} \\ \textbf{0} & 1 \end{bmatrix} \in \mathit{SE}(3)$. As shown in Fig.~\ref{FIG:1}, $\mathit{L}$, $\mathit{C}$, $\mathit{B}$ stand for the LiDAR, camera and calibration target (square board) coordinate systems respectively. The right superscript and subscript denote the two coordinate systems involved in the transformation. In our target application scenarios, the coarse extrinsic parameters ${\mathit{T}_C^L}^*$, camera intrinsic parameters and the square target length $l$ are assumed to be known. In particular, the coarse extrinsic parameter can be obtained from vehicle mechanical design. Additionally, the sensors are assumed to be time synchronized. The algorithm begins with calibration target detection from LiDAR and camera. A grid search is then performed to identify the extrinsic candidate which allows the best alignment between LiDAR and camera detection results. At last, a direct optimization method is designed to deliver the final estimation of the extrinsic parameters. Fig.~\ref{FIG:2} shows the pipeline of the proposed method.

\subsection{Camera Marker Detection}\label{CMD}
AprilTag\cite{wang2016apriltag} is widely used in robotic applications and is chosen for camera target detection in this paper. $M = \{ m_j \}, j\in\{0,1,2,3,4\}$ represents the center and the four vertices of the AprilTag pattern in the target coordinate system $B$, and $\mathit{D} = \{ d_j \}, j\in\{1, 2, 3 ,4\}$ represents the corners detected in a image. Due to AprilTag's encoding system, corners detected in the image can establish one-to-one correspondence with the target vertices. The transformation between the camera and a calibration target $\mathit{T}_B^C$ can then be estimated using the PnP algorithm. Fig.~\ref{FIG:1} shows an example of the 2D-3D correspondence. For each target, the camera detection algorithm outputs the four image corners and $\mathit{T}_B^C$. 

\subsection{LiDAR Marker Detection} \label{LMD}
This section presents our two-stage algorithm for detecting markers on LiDAR point cloud. Prior to finding potential markers, the point cloud is segmented into several clusters. Then a descriptor representing each cluster is extracted. Considering that only a small portion of clusters contain markers, a descriptor-based verification is adopted to identify board-like, rectangular clusters.
\subsubsection{Point Cloud Segmentation}
As a common first step for point cloud segmentation, ground points are labeled and then removed. Some approaches utilize RANSAC or non-linear optimization to calculate a fine ground plane model, which turn out to be time-consuming. At this point, a fast and robust range image-based method \cite{bogoslavskyi2016fast} is utilized for segmenting both ground points and non-ground points.  

Based on the vertical and horizontal resolutions, each 3D point in the LiDAR sensor's coordinate system is projected onto a cylindrical range image. Given that the scanning pattern of a MEMS LiDAR might be unevenly distributed, points that locate far in the frustum are removed before calculating the average range of a pixel. Then, with the range image calculated, a four-neighbor breadth-first search for component analysis is conducted pixel-wise\cite{bogoslavskyi2016fast}. At last, an entire point cloud is segmented into a ground cluster and other non-ground clusters, and the ground cluster is removed.

\subsubsection{Descriptor Matching}
After segmentation, we extract for each non-ground cluster a descriptor which is robust to translation and rotation variance. Existing point cloud descriptor representations mainly fall into two categories: local descriptors and global descriptors. For the sake of computational efficiency and run-time memory cost, global descriptors are considered since they describe the entire object shape and require less computations compared to the local ones. Several types of global descriptors were compared to select the most suitable one for LiDAR board detection, including M2DP\cite{he2016m2dp}, Ensemble of Shape Functions (ESF)\cite{wohlkinger2011ensemble} , Z-projection\cite{muhammad2011loop}. M2DP method presents the best performance in terms of precision-recall and time costs for our application.

Before calculating a descriptor, the principal component analysis is performed on each cluster. Then, the points of a cluster are transformed from the sensor's coordinate system to the eigen space defined by the principal axes and cluster center. In general, M2DP counts the number of point projections in the circular bins of multiple pre-defined planes whose origin aligns with the cluster center. The feature matrix is consisted of multiple rows of projections number vector, and is later decomposed by SVD to form the final descriptor with left and right singular vectors. Different from the original implementation, randomized SVD (rSVD) \cite{feng2018faster} is employed to accelerate descriptor calculation, which could reduce $50\%$ computational time of matrix decomposition. 
\begin{equation}\label{eq.PCC}
PCC = \frac{\sum_i^n(x_i-\bar x)(y_i-\bar y)}{\sqrt{\sum_i^n(x_i-\bar x)^2\sum_i^n(y_i-\bar y)^2} ~}
\end{equation}

To complete the detection procedure, a descriptor matching method is a must since different observations of a cluster will not be exactly the same. Firstly, as shown in Fig.~\ref{lidar_marker_det_sup_b}, reference descriptors are calculated based on simulated marker board cloud clusters and stored. To compare the similarity between two descriptors, M2DP \cite{he2016m2dp} simply utilizes Euclidean distance (L2 norm) as the similarity measure. Apart from Euclidean distance, other types of similarity measure for descriptor matching are evaluated, including chi-square distance and Pearson's correlation coefficient (PCC). The latter performs the best regarding robustness and precision-recall. For two n-dimensional vectors $x$ and $y$ , the Pearson correlation coefficient is defined in the equation\ref{eq.PCC}], where $\bar x$ and $\bar y$ represent mean values and gives an output value ranging from $[-1,1]$. At last, the negative correlation part is neglected in descriptor matching score, and only the clusters whose matching score exceed the chosen threshold are preserved.

\subsection{Grid Search}\label{sec:grid_search}
In practice, the translation installation accuracy is of millimeter level. However, it is common to see rotation installation error on the order of one and even a few degrees. Rotation calibration error has a large impact on LiDAR-camera data association, especially for distant objects. Thus a coarse search step is introduced to obtain a more precise initial rotation for the subsequent LiDAR extrinsic optimization.

\subsubsection{Extrinsic Perturbation}
Rotation perturbations are generated by sampling roll, pitch and yaw angles within a predefined range at fixed intervals, and each perturbed angle then is converted to a rotation matrix. Combined with initial extrinsics, a series of extrinsic candidates are generated $\mathit{PT}_C^L = \{{\mathit{T}_C^L\}_i^*}, i = 0,1,2...$

\subsubsection{Extrinsic Candidate Search}
With the target pose in camera coordinate system ($\mathit{T}_B^C$) and an extrinsic candidate in $\mathit{PT}_C^L$, the board center $^B{m_0}$ can be transformed to $^L{m_0}$ in LiDAR coordinate system for each target. Throughout the paper, the left superscript indicates the coordinate system that this variable refers to. The quality of the extrinsic candidate is judged based on the total number of LiDAR target points within a fixed-length radius around $^L{m_0}$ for all targets. With our experiments showing that the popular k-nearest neighbor(kNN) algorithm is time-consuming with a large number of image-LiDAR frame pairs and several targets in each frame, a novel range image-based search method is proposed to accelerate calculations. First of all, each point from LiDAR marker detections \ref{LMD} is projected onto a range image based on specified sensor's resolutions as the reference data structure for searching. Then, the regions for radius search in range image are calculated based on projections of camera marker detection centers. In order to find the region of interest (ROI) in range image for judging alignment, eight vertices of a bounding box whose side length equals the marker board's length and which is centered at $^L{m_0}$ are calculated first. Then, the ROI for each marker detection in range image is defined by the minimum bounding rectangle of these eight vertices' projections. At last, the number of LiDAR points in ROI is counted pixel-wise, which forms the measure of our judging standard. The best extrinsic candidate has the largest points count number. Alg. \ref{alg.l1} shows the implemented search algorithm. 

\begin{algorithm}
\caption{A pseudo code of grid search}
\label{alg.l1}
\begin{algorithmic}[table.1]
    \Require extrinsic perturbations ${PT}_C^L$, marker pose ${T}_B^C$, marker center $^B{m_0}$, marker length $l$
    \Ensure best extrinsic $\Tilde{T}_C^L$
    \Function{CoarseGridSearch}{}
    \State Perturb initial $\mathit{T}_C^L$ to get $\mathit{PT}_C^L$
    \State Project segmented LiDAR points to range image $I$
    \State Init $roi=\mathit{zeros}(4),max\_count=0,id=0$
    \For{${\mathit{T}_C^L}_i^* \in \mathit{PT}_C^L$}
    \State $^L{m_0} = \{\mathit{T}_C^L\}_i^* * \mathit{T}_B^C * ^B{m_0}$
    \State CalculateROI($^L{m_0}, l, I, roi$)
    \State Iterate over $roi$, sum points number to $count$
    \If{$max\_count < count$}
        \State $max\_count = count$
        \State $id = i$
    \EndIf
    \EndFor
    \State  $\Tilde{T}_C^L = {\mathit{T}_C^L}_{id}^*$
    \label{code:recentEnd}
    \EndFunction
    \Function{CalculateROI}{$^L{m_0}, l, I, roi$}
        \State bbox\_vertices = CalculateBBoxVertices($^L{m_0}, l$)
        \For{$vertice \in bbox\_vertices$}
            \State $\{row, col\}^I$ = RangeImageProjection($vertice, I$)
            \State $roi(0) = max(row, roi(0))$
            \State $roi(1) = max(col, roi(1))$
            \State $roi(2) = min(row, roi(2))$
            \State $roi(3) = min(col, roi(3))$
        \EndFor
    \EndFunction
\end{algorithmic}
\end{algorithm} 

\subsection{Final Optimization}\label{sec:opt}
The coarse grid search step produces initial extrinsic parameters ${\mathit{T}_C^L}^*$ as well as the LiDAR-camera target correspondences. As in \cite{huang2020improvements}, for each target, the transformation $\mathit{T}_B^L$ is first optimized using $\mathit{T}_B^C * {\mathit{T}_C^L}^*$ as the initial value. For each target point cloud cluster $\mathit{P} ={\{{}^Lx_k,{}^Ly_k,{}^Lz_k \}}_{k=1}^{N}$, the cost function $C({\mathit{T}_B^L})$ is defined as below:
\begin{equation}\label{vertextequation1}
c(\lambda, \alpha) =
\begin{cases}
min\{ |\lambda - \alpha| , |\lambda + \alpha| \}, & \text{if } |\lambda| > \alpha \\
0 & \text{otherwise }
\end{cases}
\end{equation}
\begin{equation}\label{vertextequation2}
^B{\mathit{P}} = \mathit{T}_B^L * {^L{\mathit{P}}}
\end{equation}
\begin{equation}\label{vertextequation3}
C({\mathit{T}_B^L}) = \sum_{k=1}^{N} c({}^Bx_k, l/2) + c({}^By_k, l/2) + c({}^Bz_k,\alpha)
\end{equation}

Then the target vertices in the LiDAR coordinate system is easily calculated by the transformation of $M = \{ m_j \},j=1,2,3,4$ with $\mathit{T}_B^L$. After the vertices coordinates of calibration targets in all LiDAR scans are estimated, the final extrinsic can be optimized using the $\mathit{PnP}$ and $\mathit{RANSAC}$ algorithms. This approach is termed as the indirect optimization method in this paper.

\begin{figure}[htbp]
\centering  
\subfigure[$\alpha = 0.05$ m]{
\label{Fig.boad5.1}
\includegraphics[width=7.0cm,height = 1.6cm]{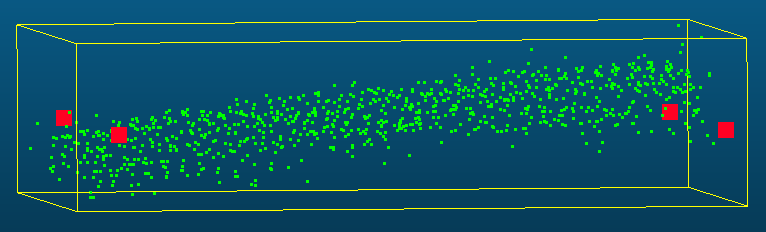}}
\qquad
\subfigure[$\alpha = 0$ m]{
\label{Fig.board.0}
\includegraphics[width=7.0cm,height = 1.6cm]{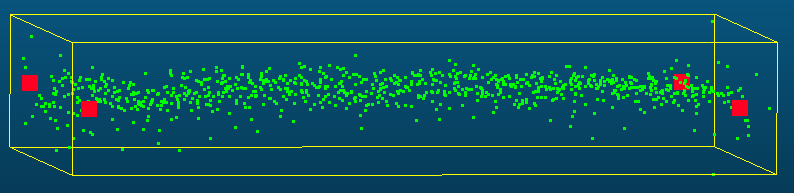}}
\caption{Green points are the segmented LiDAR cloud points projected to the target coordinate frame $B$ with the optimized $\mathit{T}_B^L$, and the red points are the vertices $M$ expressed in $B$. The two sub-figures show the case with different settings for the parameter $\alpha$.}
\label{board_proj}
\end{figure}

In \cite{huang2020improvements}, the parameter $\alpha$ in z-axis is tuned according to LiDAR measurement noise, however the effect of different parameter settings is not discussed. If the value is overestimated (larger than real noise level), it will result in insufficient constraints for $\mathit{T}_B^L$ optimization, thereby leading to inaccurate vertex estimation. Fig.~\ref{Fig.boad5.1} shows such an example where the estimated vertex does not agree with the LiDAR point cloud. The parameter tuning will be further discussed in the experimental part.

\begin{figure}[htbp]
\centering
\subfigure[Target is out of LiDAR FOV]{
\includegraphics[width=4cm, height=3cm]{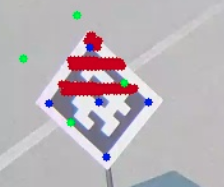}
}\subfigure[Target is too far away]{
\includegraphics[width=4cm, height=3cm]{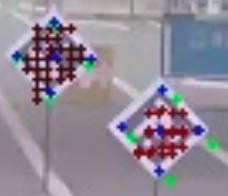}
}
\caption{Red points are the projections of target points in the image, while blue points are Apritag's detection results and green points are fitted LiDAR target vertices. (a) shows the case where the target is partially out of LiDAR's FOV when it is too close to the sensor. (b) shows the targets that are too far away from the sensor. In this case, the target point cloud becomes sparse and irregular. Single target fitted vertices in both scenarios are not precise for calibration.}
\label{wrong_data_coloect}
\end{figure}

For the indirect optimization method, the accuracy of LiDAR vertex fitting is prone to sparse and sometimes partial LiDAR observation of calibration boards. Fig.\ref{wrong_data_coloect} shows two common scenarios encountered in data acquisition: the calibration targets are either too far or too colse to the sensors. In both cases, the incomplete point cluster can not provide enough constraints to fit precise vertices. Those sub-optimal vertices can adversely affect the precision and robustness of calibration. Inspired by the well-known direct method used in visual odometry \cite{dso2018}, a direct optimization method is proposed which still uses the cost in Eq.~\ref{vertextequation1}. However, instead of optimizing $\mathit{T}_B^L$ for each individual LiDAR target, the extrinsic parameters are directly optimized using raw LiDAR board points, avoiding the intermediate vertex fitting step. For direct optimization, these incomplete point clouds combined together are able to give sufficient constraints for extrinsic optimization. Assuming $K$ target correspondences in total, the cost can be written as
\begin{equation}\label{4}
C({\mathit{T}_C^L}) = \sum_{j=1}^{K} C_{j}({\mathit{T}_B^C}*{\mathit{T}_C^L})
\end{equation}
\begin{equation}\label{5}
C_j({\mathit{T}_B^C}*{\mathit{T}_C^L}) = \sum_{k=1}^{N} c({}^Bx_k, \frac{l}{2}) + c({}^By_k, \frac{l}{2})  + c({}^Bz_k, \alpha)
\end{equation}

\section{Experiments}
The proposed algorithms are tested using real sensory data collected from our vehicle platform. The work of \cite{huang2020improvements} is chosen as the baseline, which is a state of the art method using square targets and is also most related to our work.
\subsection{Datasets}
Square calibration boards used in the experiment have a fixed length of 0.6 m and the inner AprilTag marker pattern has a length of 0.48 m. These standard foam boards are mounted on movable stands, with no additional material specifications required. Our camera has a resolution of 3840x1920 pixels. To verify the generalization performance of our proposed algorithms, Robosense\footnote{http://robosense.ai/}, Innovusion\footnote{https://www.seyond.cn/} and Hesai\footnote{https://www.hesaitech.com/} are selected for data collection. The vehicle is required to move back and forth within the range of 6 to 30 meters from the calibration targets. Each trajectory lasts around 20 seconds, including 200 LiDAR-camera frame pairs. For each dataset, 5 trajectories are included. Fig. \ref{projection_result} illustrates the data collection scenes.
\subsection{Evaluation of LiDAR Marker Detection} \label{EvalLBD}
\begin{figure}[htbp]
\centering
\includegraphics[scale=0.6]{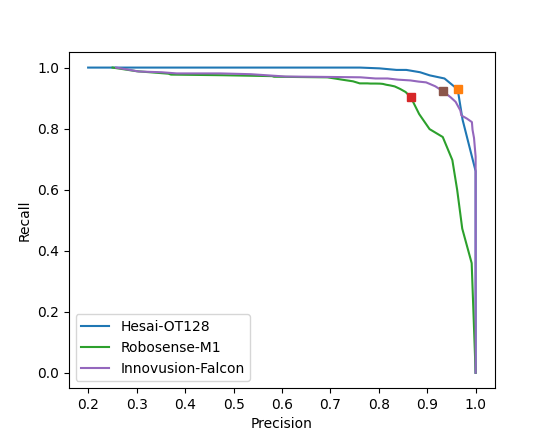}
\caption{Precision/recall of LiDAR marker detection from different types of LiDAR using our dataset.}
\label{fig.LiDAR_board_PR}
\end{figure}

As \ref{LMD} presented, we proposed a novel, general and robust LiDAR marker detection algorithm based on geometric descriptor matching without any requirement for board materials. In order to verify the versatility and robustness of the algorithm, the marker centers from different types of LiDAR data in our dataset are annotated. A detection is considered a True Positive (TP) when its center is close to a manually labeled center, otherwise a False Positive (FP). Meanwhile, a labeled center without any detection attached will be regarded as a False Negative(FN). Fig. \ref{fig.LiDAR_board_PR} illustrates curves of different types of LiDAR, including two MEMS LiDAR (Innovusion Falcon, Robosense M1) and one mechanical LiDAR (Hesai OT128), showing that our proposal can adapt to different LiDAR scanning patterns. In this figure, the descriptor matching score threshold for binary classification is set ranging from 0.7 to 0.99, and the threshold used in this paper is 0.94 and same for different types of LiDAR. The square markers in this figure correspond to a score threshold of 0.94.

\subsection{Comparison of the Direct and Indirect Optimization Method}
\begin{figure}[htbp]
\centering
\includegraphics[width=4cm]{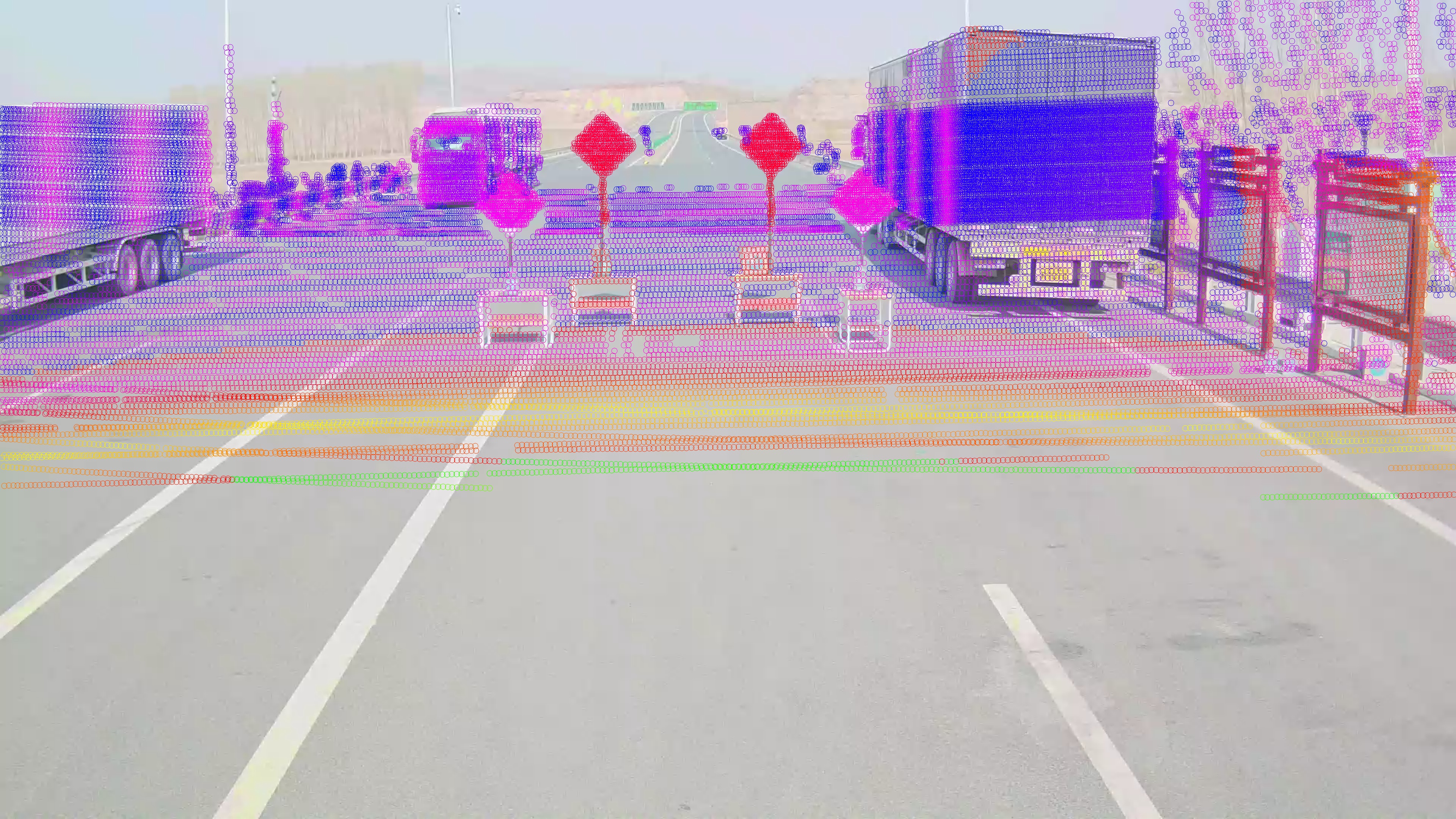}
\includegraphics[width=4cm]{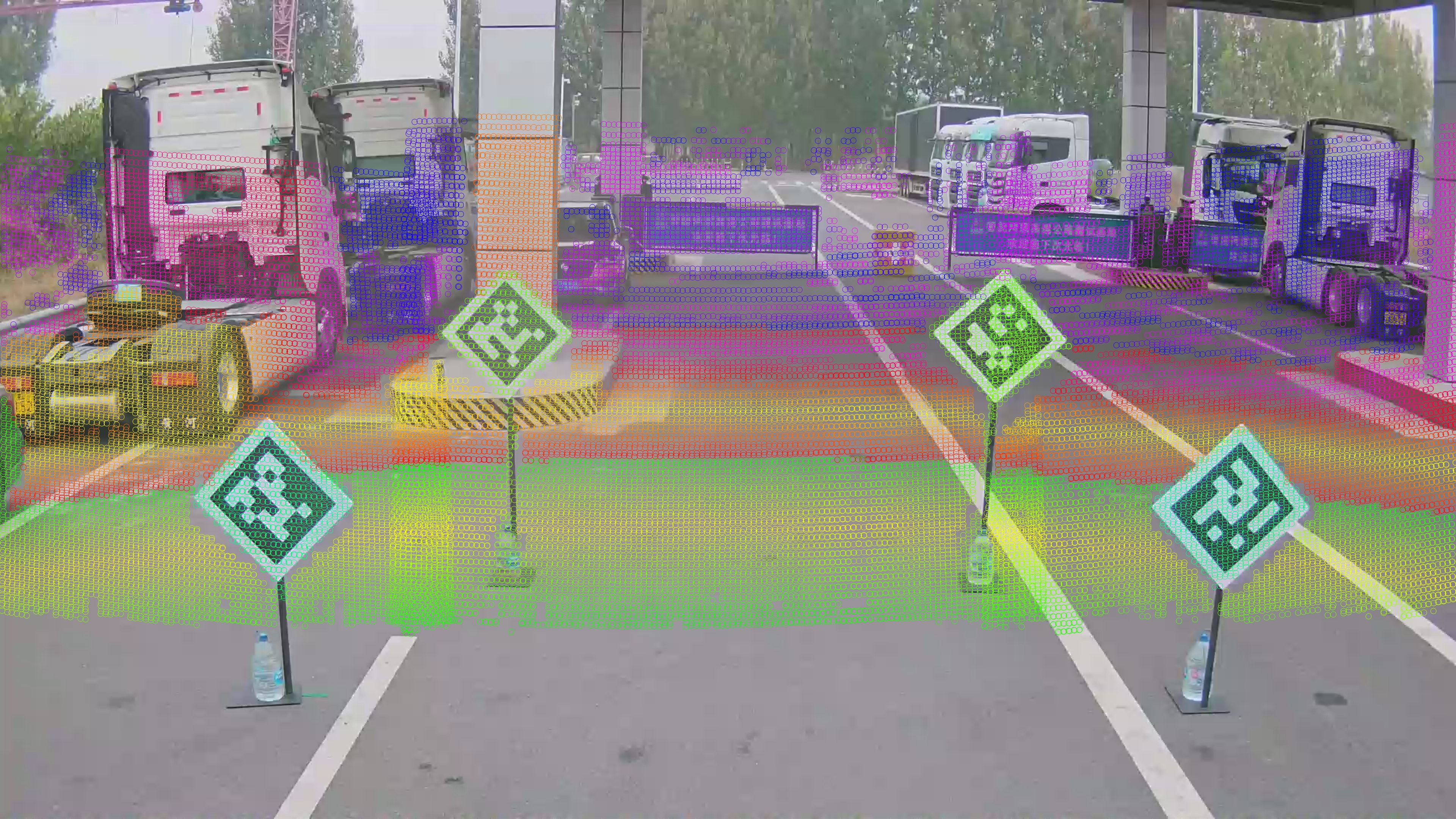}
\caption{Projection of LiDAR points onto images. The color of projection is determined by the distance of a point from the camera.}
\label{projection_result}
\vspace*{-\baselineskip}
\end{figure}

The work of \cite{huang2020improvements} did not study the influence of $\alpha$ on the calibration accuracy. Table \ref{accuracy_table} shows the average projection error (in pixels) between fitted LiDAR vertices and detected camera corners using the indirect optimization method. It is seen that $\alpha = 0$ is the best setting. It is also noted that the projection errors of Robosense are larger that the of Innovusion, caused by its larger measurement error compared to Innovusion. Fig.~\ref{projection_result} shows an example of projected LiDAR points in the image with the optimized extrinsic parameters. 

\begin{figure}[htbp]
	\centering
    \includegraphics[scale=0.18]{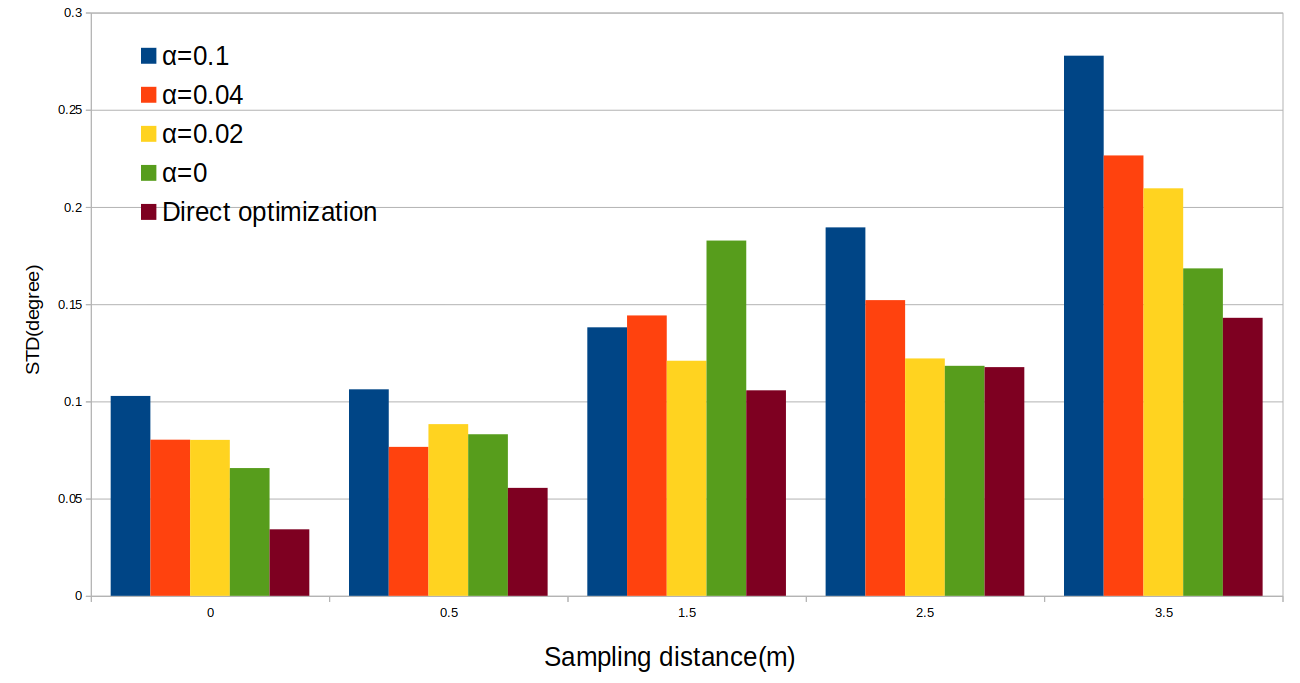}
	\caption{STD of different parameter setting and distance sampling rate. Direct optimization method exhibits a distinct advantage in robustness over the vertex method. On the other hand, as the amount of data used decreases, the depth uncertainty in vertex fitting is further manifested. }
	\label{figure_std}
\end{figure}

The collected data was down-sampled by distance to compare the robustness of the two optimization methods. That is, a data pair is selected as input for subsequent algorithms only when the vehicle has moved a certain distance. Fig.~\ref{figure_std} illustrates the standard derivations (STD) of rotational calibrations. The STD is calculated for 5 different trajectories and the average result (in degree) for roll, pitch and yaw is shown. As the sampling distance increases, the indirect method becomes much more unstable than direct method.  Table.~\ref{inproper_dist} further compares the calibration results of the two optimization methods using data collected at only close or far distances from the calibration targets, shown in Fig.\ref{wrong_data_coloect}. 
\begin{table}[htbp]
\centering
\caption{Average projection error of 5 trajectories in each dataset} 
\setlength{\tabcolsep}{3.0mm}{
\begin{threeparttable} 
\begin{tabular}{ccccc} \toprule
Parameter  &  Robosense    &  Innovusion  &  Hesai \\ \hline
$\alpha=0.1$  & 9.313  & 5.785 &  8.144 \\
$\alpha=0.06$  & 7.519 & 4.718 & 6.310 \\
$\alpha=0.05$  & 7.083  & 4.372 & 5.775 \\
$\alpha=0.04$  & 6.807  & 4.243 & 5.348 \\
$\alpha=0.03$  & 6.528  & 4.095 & 5.071 \\
$\alpha=0.02$  & 6.425  & 4.061 & 5.013 \\
$\alpha=0.01$  & \textbf{6.322} & 4.031 & 5.058 \\
$\alpha=0.0$ & 6.351  & \textbf{3.983} & \textbf{4.824}  \\
\bottomrule
\end{tabular}
\end{threeparttable} }
\label{accuracy_table}
\end{table}
\subsection{Evaluation of Coarse Grid Search}
In order to show the effectiveness of the coarse grid search, a dataset is randomly selected and rotation perturbations are added to the accurately calibrated parameters as the initial value. A rotation perturbation is generated by randomly sampling roll, pitch and yaw angles respectively in the range of -10 to 10 degrees. The search range for our grid search algorithm is from -9 to 9 degrees with the search step being 1.5 degrees. A total of 200 tests are performed, where 100 tests use the proposed grid search to provide initial calibration parameters while the remaining 100 tests directly perform extrinsic optimization. All of the tests using the grid search successfully converged. In contrast, only 6 tests succeed for the 100 tests without using grid search. 

\subsection{Algorithm Timing}
To measure the timing of the algorithm, experiments were performed on a computer equipped with an Intel(R) Core(TM) i7-10750H processor (2.60 GHz, 12 cores) and 32 GB of DDR4 memory. The processing time for 200 image and LiDAR frame pairs is around 18.7 s, 4.93 s, 0.12 s and 1.25 s respectively for LiDAR marker detection, image marker detection, grid search and direct optimization. Note that the indirect optimization algorithm takes 1.63 s. 
The results demonstrate that the algorithm proposed in this paper is highly efficient and is capable of meeting the real-time requirements of production scenarios.


\section{Conclusion}
A fully automatic approach for LiDAR-camera calibration is proposed in this paper aimed for real world applications in factory manufacturing pipeline and after-sales service. Our method can be applied to different LiDAR sensors, is able to deal with large initial extrinsic errors and achieves superior robustness and accuracy compared to existing algorithms, as demonstrated through experimental evaluations. For future work, the joint calibration of camera intrinsic parameters will be explored within the direct optimization framework.   

\begin{table}[htbp]
\centering
\caption{Two optimization methods' average STD of euler angles at close and far distance} 
\setlength{\tabcolsep}{3.0mm}{
\begin{threeparttable} 
\begin{tabular}{c|cccc} \toprule
Datasets & \multicolumn{2}{c}{Close Distance\tnote{1}} & \multicolumn{2}{c}{Far Distance\tnote{2}} \\
Methods  &  Indirect  &  Direct  &  Indirect  &  Direct \\ \hline
$roll \quad std$  & 0.840 & \textbf{0.213} &  0.304 & \textbf{0.215} \\
$pitch \quad std$  & 0.165 & \textbf{0.159} & 0.802 & \textbf{0.307} \\
$yaw \quad std$  & 0.437 & \textbf{0.220} & \textbf{0.370} & 0.372 \\
\bottomrule
\end{tabular}
\begin{tablenotes}
\footnotesize
\item[1] Within 8.5m between targets and sensors 
\item[2] Above 25m between targets and sensors 
\end{tablenotes} 
\end{threeparttable}}
\label{inproper_dist}
\end{table}

\vspace*{-\baselineskip}

\bibliographystyle{IEEEtran}
\bibliography{Lidar-camera}
\vspace{12pt}

\end{document}